This document describes the research project for the CIFRE PhD defined in collaboration between IBM France Lab and Ecole Centrale Paris (Laboratoire Génie Industriel)

**Solution Repair/Recovery in Uncertain Optimization Environment**

PhD Candidate: Oumaima Khaled
IBM PhD Supervisor: Xavier Ceugniet
Lab PhD Supervisors: Michel Minoux, Vincent Mousseau

## Introduction

Operation Research deals with the application of advanced analytical methods such as optimization models applied to help in taking better decisions. Theoretically, models are developed, tested and then operated in an organization. Unfortunately, the environment wherein the solution is developed differs from the one wherein the solution is operated. Different unexpected events occur while operating the optimal decision, therefore there is a need expressed by the operators which is the ability to quickly adapt a plan in operation to take into account events making it anymore or in part inapplicable.

During the last few years, Optimization Solution Team within IBM has been involved in the development of optimization solutions for planning and scheduling for industrial markets with strong potential. To better match the needs of their customers, they seek to develop methods and software tools for repair or recovery.

The current research project is intended as a contribution to equip the existing optimization models of IBM (mainly Mixed-Integer Programming, Linear Programming or Constraint Programming) with repair capabilities in order to propose optimization solutions which can be operated as widely as possible.

## Context

Operation management problems (such as Production Planning and Scheduling) are represented and formulated as optimization models. The resolution of such optimization models leads to solutions which have to be operated in an organization. However, the conditions under which the optimal solution is obtained rarely correspond exactly to the conditions under which the solution will be operated in the organization.

Therefore, in most practical contexts, the computed optimal solution is not anymore optimal under the conditions in which it is operated. Indeed, it can be "far from optimal" or even not feasible. For different reasons, we hadn't the possibility to completely re-optimize the existing solution or plan. As a consequence, it is necessary to look for "repair solutions", i.e., solutions that have a good behavior with respect to possible scenarios, or with respect to uncertainty of the parameters of the model. To tackle the problem, the computed solution should be such that it is possible to "repair" it through a local re-optimization guided by the user or through a limited change aiming at minimizing the impact of taking into consideration the scenarios.

## Goals

As announced in the introduction, the operational purpose of this research project is the development of repair/recovery infrastructures which will be added to the existing solutions developed by IBM. In view of achieving this goal, we propose to carry out some preparatory research work concerning the optimization of two problems related to production planning and scheduling.

The first step in this investigation will be to advance the state of the art in the models and algorithms used to formulate and solve the problems. It tries to optimize problems related to production planning and to scheduling in order to propose mathematical formulations for the existing solutions. Therefore, we will have the possibility to compare them with models already proposed in the literature and to evaluate to what extent the solutions proposed existing approaches are different.

After modeling the existing solutions and validating them using Cplex and OPL, the next objective of this research will be to calculate a repair solution given:

- A current plan with a common set of decision variables
- A number of scenarios grouping a set of frozen variables and relaxable constraints
- A new objective function composed of a combination of a set of Key Performance Indicator (KPI) presenting the impact of taking into consideration scenarios and the initial solution.

The repair solutions will be built on existing functionalities developed by IBM such as Complex Project Scheduling (CPS) and Production Planning and Scheduling (PPS). It will be conducted in relation with the development of infrastructure configuration solutions, allowing business experts to configure 'repairable' solutions.

## Research Organization

We will start by investigating robust optimal solution and concepts for solution repair. We will deal with two study cases belonging to the family of CPS and PPS and try in an early stage to extend these solutions and concepts to more general optimization problems.

These study cases are developed in a platform called IBM ILOG ODME (Operational Decision Management Enterprise). ODME is a platform to implement and deploy corporate custom solutions for optimization-based planning and scheduling. Developing a realistic plan or schedule provides the best possible balance between customer service and revenue goals. With ILOG ODM Enterprise, business leaders can make better decisions through what-if analysis, scenario management, and collaboration.

### 1. Tail Assignment

The first study case is a solution developed by IBM in the context of airline operations called Tail Assignment which is the problem of creating routes for a set of individual aircraft, covering a set of flights in a timetable so that various operational constraints are satisfied while minimizing some cost functions.

Tail Assignment or Aircraft routing is a part of the sequential procedure used by airlines to plan their operations. They first start by modeling the market behavior to construct a timetable containing information about the time and place of the arrival and departure of each flight. Then, the fleet assignment is performed to assign an aircraft type to each flight leg to maximize anticipated profit while taking into account the number of available aircraft. For each aircraft type, an aircraft routing problem is then solved by determining the sequence of flight legs to be flown by each individual aircraft.

Before investigating the issue of repair, we will start formulating Tail assignment as a MIP mathematical model [Minoux M. 2008]. The model consists as described above of assigning an aircraft to a flight while respecting the main constraints below:

**C1** : Minimum allowed time between the arrival and departure of an aircraft, needed for activities of cleaning, refueling, changing crew and passengers….

**C2** : Planes must depart from an initial position, that means the first flight operated by an aircraft must have as a departure airport the initial position of the aircraft

**C3** : Flights are with no ferry represent a flight without passengers or even commercial flights. For one aircraft, the arrival airport for a flight should be the same as the departure flight operated by the same aircraft.

Generally, modeling and solving aircraft routing problem are well studied in the literature [Liang Z., Chaovalitwongse W. A. 2009] and recently in regard to the occurrence of disruptions such as flight delays, cancellation... there has been a focus on introducing robustness into airline planning phases to mitigate the effect of these disruptions.

Normally strict robustness [Bertsimas D., Sim M. 2003] [Ben Tal A., Nemirovski A. 2002] requires a planned solution to perform under all disruptions, but what we need is to recognize that in a disruption, repair also called in literature 'recovery' is required and the planned solution will be changed in operations. The contrast between robust and repair is well explained in the article of [Liebchen C. et al. 2009].

To summarize, we will develop a two stage Tail assignment model composed by a planning stage and a recovery stage based on specific scenarios proposed by IBM such as Flight delay, cancellation, and last time aircraft unavailability [Minoux M. 2011]... The two stages are solved either simultaneously or separately to improve the robustness of the planned solution.

**2. Production Planning**

The second study case is a classical production planning problem [Murla J. et al. 2006]. The objective of this application is to decide how much of which product must be produced at each location for a given time period.

The planning can also include the decisions on the transportation of raw materials between suppliers and production locations, transportation of intermediate products between different production locations, and transportation of finished products between production location and customers.

Objectives may include minimization of production, inventory and transportation cost along with minimization of penalties related to target inventories, or target demand and/or forecast to cover.

We will adapt the same methodology as for the first study case; starting from mathematical modeling of the problem until generating algorithm of repair based on specific scenarios. Some typical repair scenarios in this context are the following:

**Scenario 1.** A constraint requires that all customers of priority must have all their orders being fulfilled. The optimization engine is run and finds feasible and optimal solutions. The business user modifies the priority of a given customer to be greater than the threshold. This is an infeasibility that needs to be detected and repaired, without re-running the complete optimization algorithm that can take minutes or hours to run.

**Scenario 2.** A new order is created after optimization is run so that the plan does not include the decision to produce and transport products to fulfill this order. We want to detect and repair this.

**3. Generalization**

Dealing with two study cases with different models of repair helps us to study the similarity between them in order to propose a more general framework of repair which could be coupled with that of any model of mathematical programming offering to the users a better apprehension of the solutions. The general framework leans on a specific definition of scenarios, related criteria in order to define the impact of taking into consideration scenarios on the initial solution and finally repair algorithms. To model repair algorithms, a possible idea would be e.g. to explore possible connections with VNS (Variable Neighborhood Search) [Hansen P. et al. 2010]. The reason for this is that VNS is also based on repeatedly solving local reoptimization sub problems.

After modeling general algorithm of repair, we will focus on studying robustness issues and try to setup new criteria of robustness which are based on repair. Relating robustness and recovery is a fairly new research subject aiming at defining robust solutions with a minimum possible effort [Froyland G. et al. 2013]. We will explore this issue in our research in order to highlight the relationship between a robust solution and reparable solution and see how far we can conclude that reparability is a quality of robust solution that contributes to define new criteria of robustness. So we can assume that the final deliverable of our project is the elaboration of a recoverable robustness framework leading to solutions that can be repaired with limited changes.

Another subject that will be considered in our research is the relationship with the decision-maker about the criteria, the scenario set and the definition of robustness [Thengvall B.G. 2000]. There is a parallel relation with Multi Criteria Decision Making [Bouyssou D. et al. 2006] based on introducing the preference of the user. It goes toward decision aid systems leading to some degree of robustness [Greco S. et al. 2008].

To validate the general framework, we will test it in connection with other application examples provided by IBM in order to specify further integration framework in the existing optimization tools developed by IBM.

Another interesting issue that can lead to further developments of the proposed research concerns the extension of the repair algorithms to Constraint Programming models. Even if it impossible, in the present stage, to anticipate how far it will be possible to go in this direction, attempts will be made to properly identify those concepts resulting from the work which are likely to be applicable to the Constraint Programming field.

To have clearer vision of how our work will be organized, we propose the following Gantt chart of activities:

('To' denotes the starting date of the PhD)

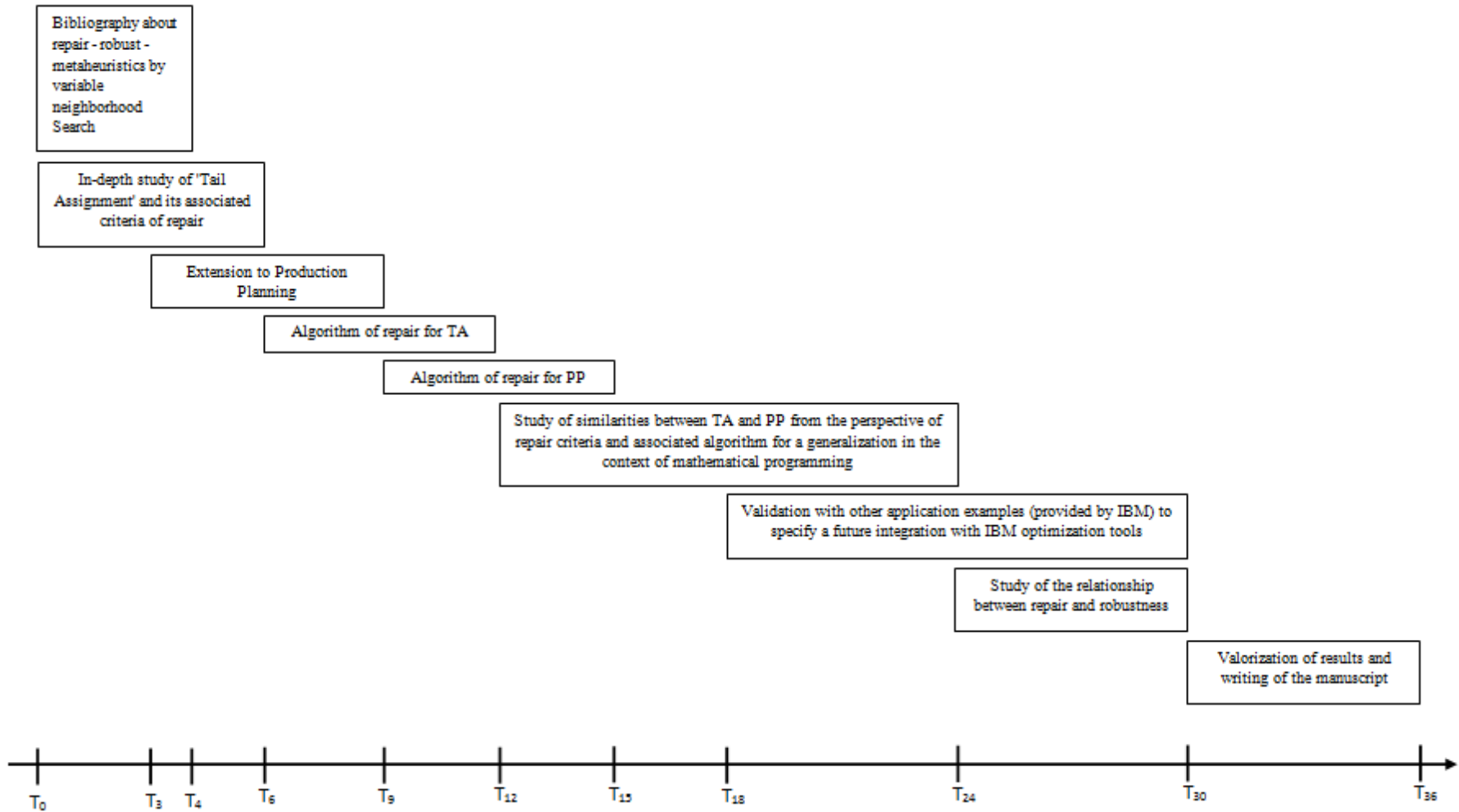

**Figure 1 : PhD organization**